\documentclass[aps,pre,twocolumn,groupedaddress,floatfix,superscriptaddress]{revtex4}
\usepackage{dcolumn}
\usepackage{amsmath}
\usepackage{amssymb}
\usepackage{lipsum}
\usepackage{graphicx}
\usepackage{bm}
\usepackage[T1]{fontenc}
\usepackage{color}
\usepackage{url}
\usepackage[bf]{subfigure}
\usepackage{rotating}
\usepackage{soul}
\usepackage{scalefnt}
\usepackage{hyperref}
\usepackage{scalefnt}
\usepackage{multirow}

\usepackage{perpage} %the perpage package
\MakePerPage{footnote}

\usepackage{color}
\definecolor{redcolor}{rgb}{1.0,0.,0.}

\usepackage[utf8]{inputenc}
\begin{document}
%%Complexity does not have a LATEX template: "Regardless of the source of the word-processing tool, only electronic PDF (.pdf) or Word (.doc, .docx, .rtf) files can be submitted" https://www.hindawi.com/journals/complexity/guidelines/
\title{An Image Analysis Approach to the Calligraphy of Books}

\author{Henrique Ferraz de Arruda}
\affiliation{Institute of Mathematics and Computer Science, University of São Paulo, Brazil}
\author{Vanessa Queiroz Marinho}
\affiliation{Institute of Mathematics and Computer Science, University of São Paulo, Brazil}
\author{Thales Sinelli Lima}
\affiliation{Institute of Mathematics and Computer Science, University of São Paulo, Brazil}
\author{Diego Raphael Amancio}
\affiliation{Institute of Mathematics and Computer Science, University of São Paulo, Brazil}
\author{Luciano da Fontoura Costa}
\affiliation{São Carlos Institute of Physics, University of São Paulo, Brazil}

\begin{abstract}
Text network analysis has received increasing attention as a consequence of its wide range of applications.  In this work, we extend a previous work founded on the study of topological features of mesoscopic networks.  Here, the geometrical properties of visualized networks are quantified in terms of several image analysis techniques and used as subsidies for authorship attribution.  It was found that the visual features account for performance similar to that achieved by using topological measurements.  In addition, the combination of these two types of features improved the performance.
\end{abstract}
\maketitle

\setcounter{secnumdepth}{1}

\section{Introduction}

With the ever increasing availability of machine readable text, the interest in automatic textual analysis tools has grown substantially given its potential for several important applications. One area of special interest is authorship attribution, which has been studied from different perspectives. In authorship attribution, we are interested in assigning an author to a given text~\cite{Stamatatos,Juola:2006}. Traditional approaches use statistical analysis of lexical and syntactical features~\cite{Stamatatos,grieve2007,Koppel:2009}, while other methods focus on text modeling using complex networks~\cite{Lahiri,Marinho2016BRACIS,1742-5468-2015-3-P03005,segarra2015authorship,interplay,MEHRI20122429}. The latter is capable of representing structural as well as semantic characteristics~\cite{Cancho2004patterns,semantic,10.1371/journal.pone.0067310,0295-5075-98-1-18002}, complementing traditional methods. 

Recently, a network method for text modeling was proposed~\cite{de2017mesoscopic}.  This method uses mesoscopic networks to capture the flow of narratives.  In ~\cite{marinho2017Calligraphy}, this approach was used for authorship attribution.  By \emph{mesoscopic} it is meant that the derived networks are able to reflect text relationships at a topological scale larger than usually approached by using, for instance, word adjacency.  It has been suggested that such model can provide insights about the discursive structure of texts~\cite{marinho2017Calligraphy}, an issue which is explored further in the current work.  

\begin{figure}[!htpb]
  \centering
    \includegraphics[width=.35\textwidth]{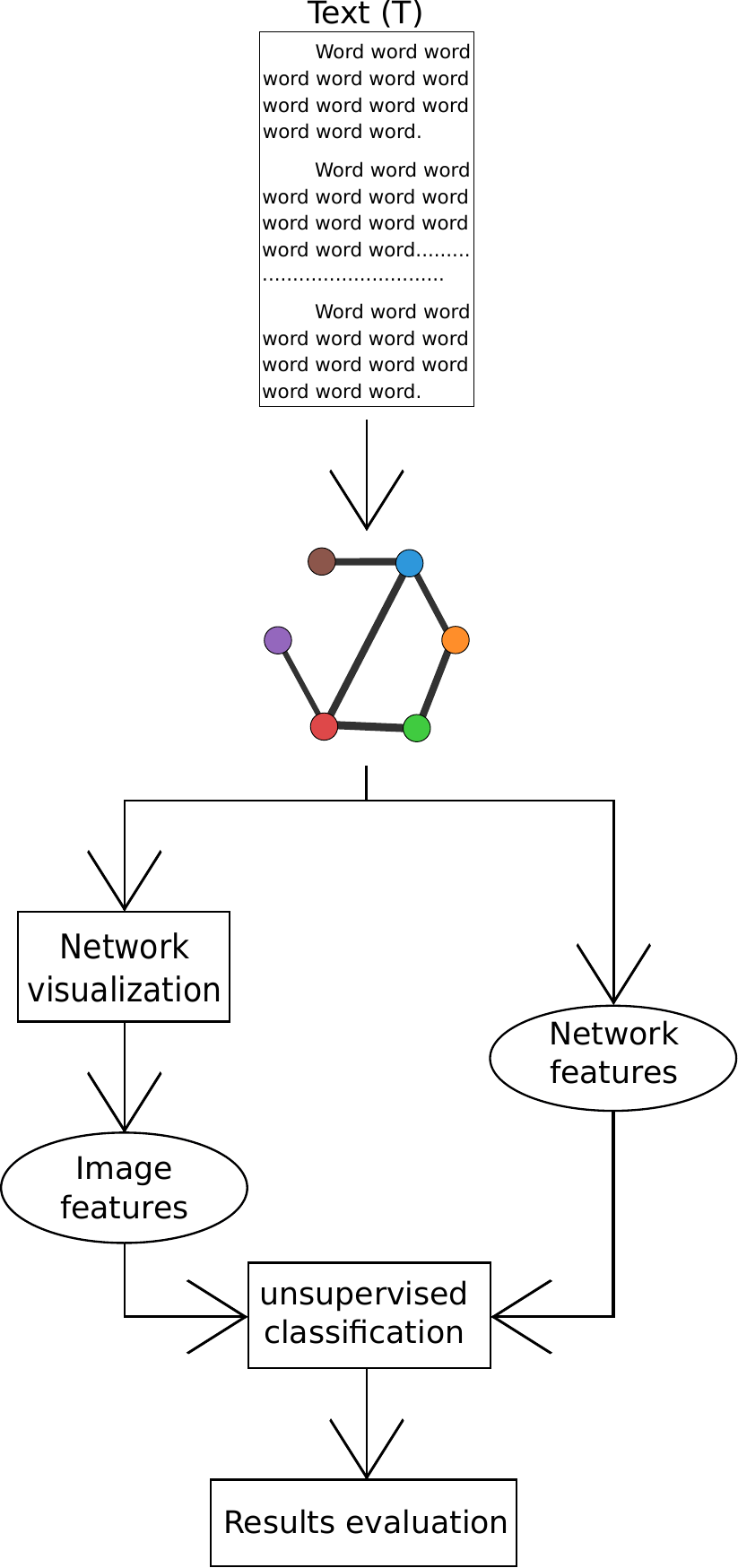}
   \caption{Scheme illustrating the mapping of texts into networks and the classification process.
   Initially, documents are mapped into networks. Then, two types of features are calculated: network features and attributes derived from network visualizations. The classification can then be performed using both network and image features.}
  \label{fig:globalschemes}
\end{figure}

The work presented in~\cite{marinho2017Calligraphy} focuses on network feature extraction, while the present incorporates features from the visualizations of these networks.  This has been done to explore how the visual features of the networks, which was called calligraphy in~\cite{marinho2017Calligraphy}, can contribute to the accuracy of authorship attribution. The framework used in this work is illustrated in Figure \ref{fig:globalschemes}. The obtained results suggest that these network visual features are as capable as topological measurements to characterize authors' styles.  Furthermore, when the visual and topological features were combined, a better result is achieved.

The remaining of this paper is organized as follows: Section~\ref{sec:related_work} defines the authorship attribution problem and presents some traditional and network based approaches. The proposed pipeline, which includes the process of obtaining mesoscopic networks and the selected image and network measurements, is presented in Section~\ref{sec:methods}. Then, Section~\ref{sec:results} reports the obtained results using image and network features. Finally, our conclusions are drawn in Section~\ref{sec:conclusions}.

\section{Related Work}\label{sec:related_work}
Given a text of unknown or disputed authorship and a set of candidate authors, the goal of an authorship attribution is to identify the correct unknown author~\cite{Stamatatos}. One of the most important findings in authorship studies was reported by Mosteller and Wallace~\cite{Mosteller}. They investigated the authorship of several political documents and discovered that the frequencies of common words -- such as some pronouns, prepositions, and articles -- are useful to characterize the authorship of texts. 

Traditionally, works in authorship attribution use simple, yet useful features to characterize writing styles, such as statistics extracted from words and characters (e.g. frequency of the $n$-grams of words or characters, and the frequency of punctuation marks)~\cite{Stamatatos,grieve2007, Koppel:2009}. Syntactic and semantic features can also be used for the task, such as the frequency of the constituent parts of a sentence and information about words synonyms~\cite{Stamatatos}. In recent years, authorship attribution has been studied from a different perspective. Some works have taken advantage of dense and real-valued vectorial representations and deep neural networks~\cite{bagnall:2016,SolorioEACL,SariEACL}, such as their application to train language models for each author or to learn continuous vectors for $n$-grams of words. 

Complex network-based approaches have also been used to assign authorship. In most of these approaches, co-occurrence networks are created from the texts~\cite{Lahiri,Marinho2016BRACIS,1742-5468-2015-3-P03005,segarra2015authorship,interplay,MEHRI20122429}. 
Basically, in such approaches, co-occurrence networks are created by connecting adjacent words. Then, several topological measurements -- such as the clustering coefficient, degree, frequency of motifs -- are extracted and their respective values are used as features in machine learning algorithms. These previous works have shown that structure plays a prominent role in characterizing authors' styles. Interestingly, even if only stopwords are used, the structure of the remaining structure in the networks is still essential for the accurate identification of styles~\cite{segarra2015authorship}. 

Another possibility for text representation is to map texts as mesoscopic networks, thus providing a model for identifying the relationship between larger chunks of texts~\cite{de2017mesoscopic}.  Interestingly, Marinho \emph{et al.}~\cite{marinho2017Calligraphy} presented mesoscopic networks created from a dataset and the provided visualizations revealed characteristics of the authors, such as the preference for short stories over novels, as well as characteristics of each text, such as the similarities between the beginning and ending of a particular book. These mesoscopic networks also characterize the unfolding of the stories.  

\section{Materials and Methods}\label{sec:methods}

\begin{figure}
  \centering
    \includegraphics[width=.45\textwidth]{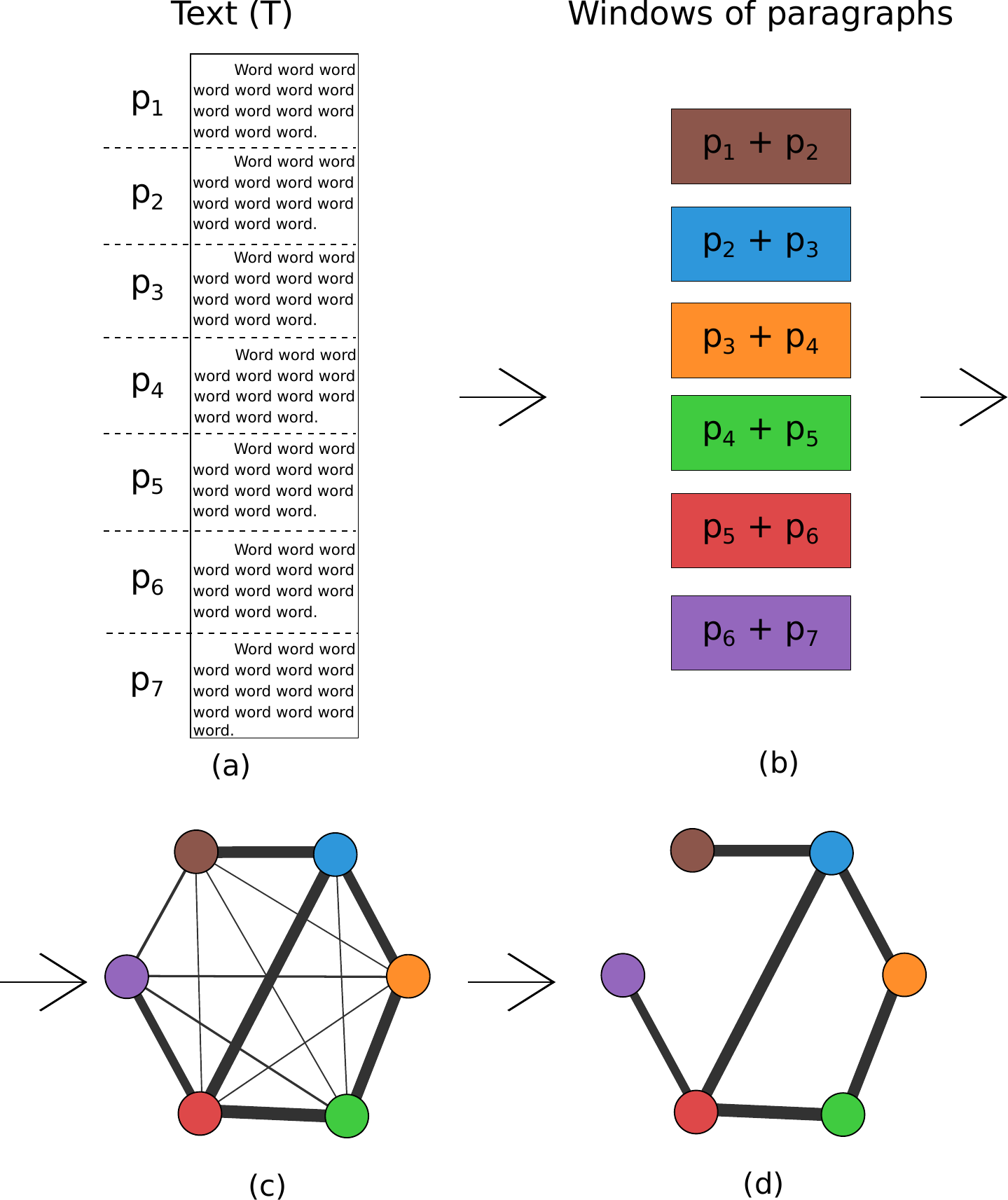}
   \caption{Mesoscopic network. In (a), the document is divided into paragraphs. The nodes of the network are defined as adjacents paragraphs in (b). The links are established according to the textual similarity of the nodes in (c). A threshold is applied in (d) so as to remove the weakest links.}
  \label{fig:mesoscopic}
\end{figure}

Our authorship attribution method is based on the mesoscopic modeling approach~\cite{de2017mesoscopic}.  The first step to create mesoscopic networks is the preprocessing of the text. In this study, we remove stopwords (such as articles and prepositions), and the remaining words are lemmatized. The preprocessed text is then partitioned into a set of paragraphs $T = \{p_1,p_2\ldots,p_n\}$, where $n$ is the total number of paragraphs, and $p_i$ is the set of words belonging to the same paragraph (see Figure~\ref{fig:mesoscopic}(a)). Taking into account the paragraphs' order, all possible sets of $\Delta$ consecutive paragraphs are grouped, as shown in Figure~\ref{fig:mesoscopic}(b). Each set $W_i^\Delta=\{p_i,p_{i+1}\ldots,p_{i+\Delta - 1}\}$ represents a network node. From these nodes, a weighted and undirected network is created, in which all nodes $i$ and $j$ are connected and the weights are computed as the cosine similarity between the vector containing the values of tf-idf statistics associated to each node~\cite{Manning:1999}, as illustrated in Figure~\ref{fig:mesoscopic}(c).  In the current study, we employed the following tf-idf mapping:
\begin{equation}
  \text{tf-idf}(w,d,D) = \frac{f_{w,d}}{n} \times \log\Bigg{(} \frac{|D|}{d_w}\Bigg{)},
\end{equation}
where $f_{w,d}$ is the frequency of a given word $w$ occurring in document $d$ comprising $n$ words, $|D|$ is the total number of documents  and $d_w$ is the number of documents in which the word $w$ occurs. Note that, in this approach, each set of consecutive paragraphs $W$, i.e. each node, is considered a document. 

%\subsection{Network pruning}
In order to obtain unweighted networks, we remove the edges with the lowest weights until a certain criteria is met, as can be seen in Figure~\ref{fig:mesoscopic}(d). In this paper, the selected criterion requires that all networks have the same network average degree $\left\langle k \right \rangle = 2E/N$, where $E$ is the number of edges and $N$ is the number of nodes. The selected value for $\left\langle k \right \rangle$ was 40. Note that such an approach was previously used in~\cite{marinho2017Calligraphy}.

\subsection{Dataset}
We used a dataset of 50 English texts written by 10 authors. These books are available from the Project Gutenberg repository\footnote{Project Gutenberg - https://www.gutenberg.org/}. The list with the 50 texts is presented in Table~\ref{tab:dataset}.

\begin{table}
\caption{List of 50 texts used in this work. The list comprises books from ten authors.}
\label{tab:dataset}
\centering
\begin{tabular}{p{8cm}}
 \textbf{Author: Texts}\\
\hline
\textbf{Herman Melville:} Moby Dick, Or, The Whale; The Confidence-Man: His Masquerade; The Piazza Tales; Typee: A Romance of the South Seas; White Jacket, Or, The World on a Man-of-War\\
\hline
\textbf{B. M. Bower:} Cabin Fever; Lonesome Land; The Long Shadow; The Lookout Man; The Trail of the White Mule\\
\hline
\textbf{Jane Austen:} Emma; Mansfield Park; Persuasion; Pride and Prejudice; Sense and Sensibility\\
\hline
\textbf{Mark Twain:} A Connecticut Yankee in King Arthur's Court; Adventures of Huckleberry Finn; The Adventures of Tom Sawyer; The Prince and the Pauper; Roughing It\\
\hline
\textbf{Charles Darwin:} Coral Reefs; Geological Observations on South America; The Different Forms of Flowers on Plants of the Same Species; The Expression of the Emotions in Man and Animals; Volcanic Islands\\
\hline
\textbf{Charles Dickens:} American Notes; A Tale of Two Cities; Barnaby Rudge: A Tale of the Riots of Eighty; Great Expectations; Hard Times\\
\hline
\textbf{Edgar Allan Poe:} The Works of Edgar Allan Poe (Volume 1 - 5)\\
\hline
\textbf{Hector H. Munro (Saki):} Beasts and Super Beasts; The Chronicles of Clovis; The Toys of Peace; The Unbearable Bassington; When William Came\\
\hline
\textbf{Thomas Hardy:} A Changed Man and Other Tales; A Pair of Blue Eyes; Far from the Madding Crowd; Jude the Obscure; The Hand of Ethelberta\\
\hline
\textbf{Henry James:} The Ambassadors; The American; The Portrait of a Lady - Volume 1; The Real Thing and Other Tales; The Turn of the Screw\\
\hline
\end{tabular}
\end{table}

\subsection{Network analysis}
\label{sec:networkAnalysis}
After the networks have been created, topological measurements can be extracted. Here, we measure accessibility (for $h = \{2,3\}$)~\cite{Travencolo2008}, degree~\cite{costa2007characterization}, backbone and merged symmetry (for $h = \{2,3,4\}$)~\cite{Silva}, assortativity~\cite{newman2003mixing}, average degree of neighbors~\cite{pastor2001dynamical} and clustering coefficient~\cite{costa2007characterization}. Because most of the selected measurements apply to a single node, some statistics -- such as the average, standard deviation, and skewness -- were extracted from each distribution and used as features for the classification algorithms. 

\subsection{Image analysis}
\label{sec:imageAnalysis}
According to Marinho \emph{et al.}~\cite{marinho2017Calligraphy}, the visualization of mesoscopic networks can provide information about the writing style of authors. In the present study, we used image processing analysis to extract characteristics from these visualizations, which are used to characterize authors. First, the complex networks are visualized, using a force directed algorithm, which is based on attraction force between connected nodes and repulsion force between all pairs of nodes~\cite{fruchterman1991graph}. To that end, we employed the software provided by Silva \emph{et al.}~\cite{silva2016using} with fixed parameters. 2D visualizations were used to remove the need to define the projection angle. Furthermore, the images were converted to monochromatic versions (called binary images). A preprocessing step was employed in which the images are dilated and eroded~\cite{delacao} in order to remove small holes. We considered the object size and the higher eigenvalue ($\lambda_1$) of the the Principal Component Analysis (PCA)~\cite{jolliffe2002principal} to find the dilation kernel as follows:
\begin{equation}
 c_{dim} = c\lambda_1,
\end{equation}
where $c$ is a constant, set as $c=3\cdot10^{-4}$. In other words, the larger the object is, the larger the kernel size. 

The following features were extracted from the preprocessed images:

\begin{enumerate}

\item \emph{\textbf{Area}}: The object area $A$ is calculated as the sum of all pixels of the object in the image (see Figure~\ref{fig:features1}); 

\item \emph{\textbf{Perimeter}}: The perimeter of an object, $P$, is the arc length of the external border of the object. An example is shown in Figure~\ref{fig:features1};

\item \emph{\textbf{Euler number}}: This measurement is derived from the number of holes in the image.  It is computed as:
\begin{equation}
e = N_o - N_h,
\end{equation}
where $N_h$ is the number of holes in the image and $N_o$ is the number of objects. An example is shown in Figure~\ref{fig:features1}, where $N_o = 1$, $N_h = 2$ and $e=-1$.  Note that for all images in this study, $N_o = 1$;

\item \emph{\textbf{Minimum enclosing circle}}: The minimum enclosing circle $M_c$ is the smallest circle that includes all of the objects' pixels~\cite{SKYUM1991121}. We measured the radius and the $x$ and $y$ coordinates of the central point of this circle.  Figure~\ref{fig:features1} illustrates these measurements;

\item \emph{\textbf{Convex hull}}: similar to the minimum enclosing circle, the convex hull consists of the smallest convex polygon including all the pixels from the object~\cite{Sklansky} (see Figure~\ref{fig:features1}).  We compute the polygons' area $C_a$, perimeter $C_p$, and residual area $C_r$.  The latter is calculated as: 
\begin{equation}
C_r = C_a - A.
\end{equation}

\begin{figure}[!htpb]
  \centering
    \includegraphics[width=0.4\textwidth]{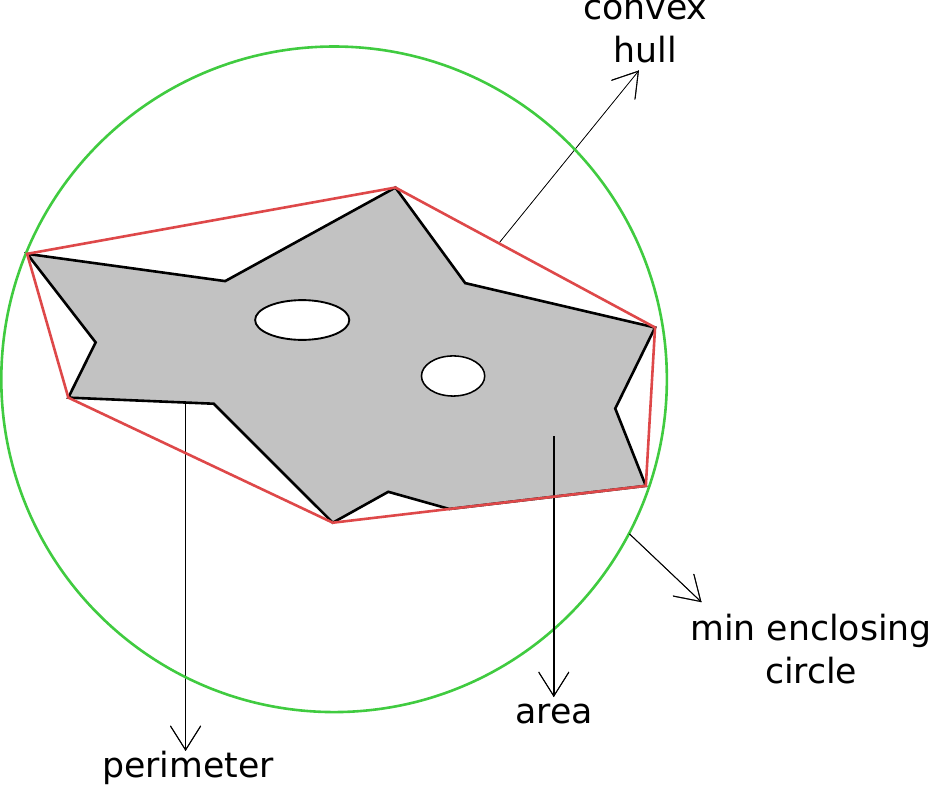}
   \caption{Example of the following image features: area, perimeter, minimum enclosing circle, and convex hull. Note that the Euler number of this image is $e = -1$ because $N_o = 1$ and $N_h = 2$.}
  \label{fig:features1}
\end{figure}

\item \emph{\textbf{Elongation}}: This measurement quantifies how stretched an object is. It is calculated as the ratio between the variances of the first ($\lambda_1$) and second ($\lambda_2$) axes of the PCA of the image~\cite{costa2000shape}.

\begin{figure}[!htpb]
  \centering
    \includegraphics[width=0.4\textwidth]{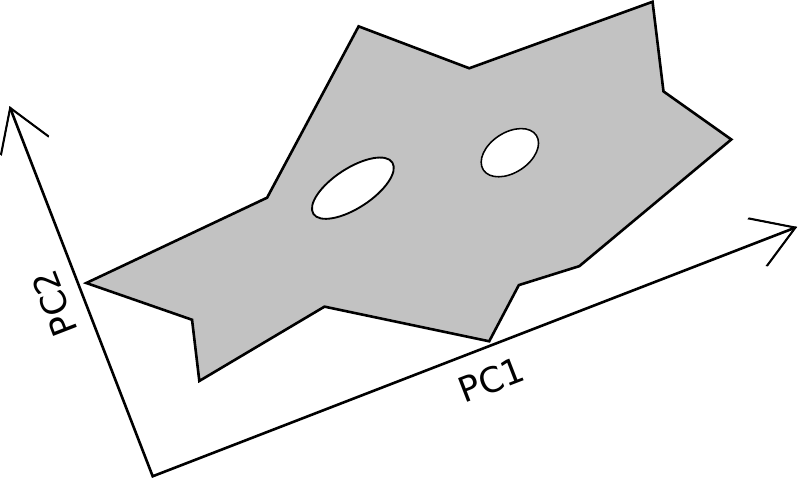}
   \caption{Example of elongation measurement, where   and $\lambda_2$ are associated to PC1 and PC2, respectively. In this example $\lambda_1 = 100$ and $\lambda_2 = 50$, so the elongation is 2.}
  \label{fig:elong}
\end{figure}

\item \emph{\textbf{Lacunarity}}: This feature~\cite{Plotnick1993} quantifies the variation of the size of the `voids' of the objects in an image, as shown in Figure~\ref{fig:lac}.  Note that when the holes are more uniform, the lacunarity ($L$) value is lower. In this study, we employed the self-referred approach to lacunarity proposed by Rodrigues \emph{et al.}~\cite{PhysRevE.72.016707}, which involves circular windows of radius $r$.  

\begin{figure}[!htpb]
  \centering
    \includegraphics[width=0.4\textwidth]{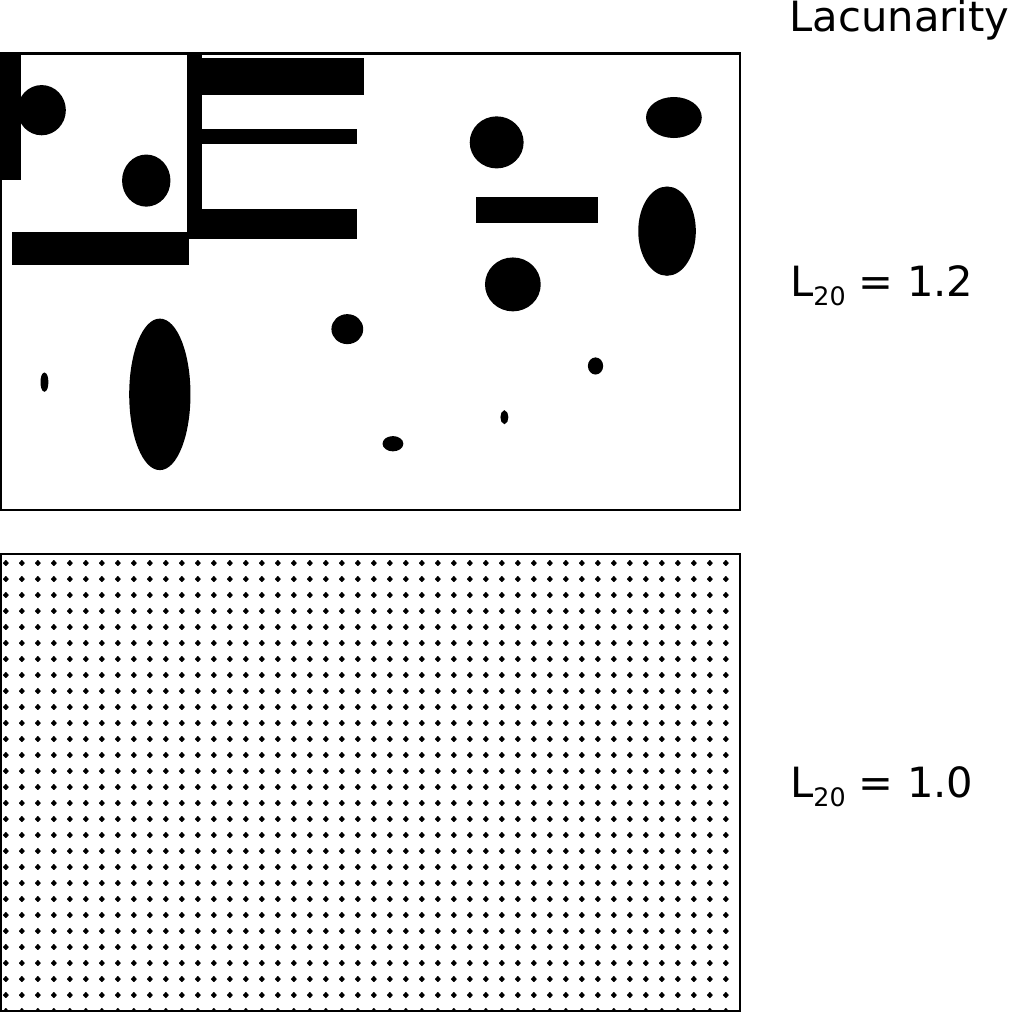}
   \caption{Two examples of images and their respective lacunarity with $r=20$. Here, the objects are shown in black, and the holes are white.}
  \label{fig:lac}
\end{figure}

\item \emph{\textbf{Fourier}}: The Fourier transform represents the signal (image) in the frequency domain~\cite{costa2000shape}. In this paper, we consider the magnitude of this transform. Non-overlapping rings with fixed width are defined in the magnitude Fourier space. The entropy, average, and standard deviation value of magnitudes inside each ring are obtained and employed as features. 

\begin{figure}[!htpb]
  \centering
    \includegraphics[width=0.45\textwidth]{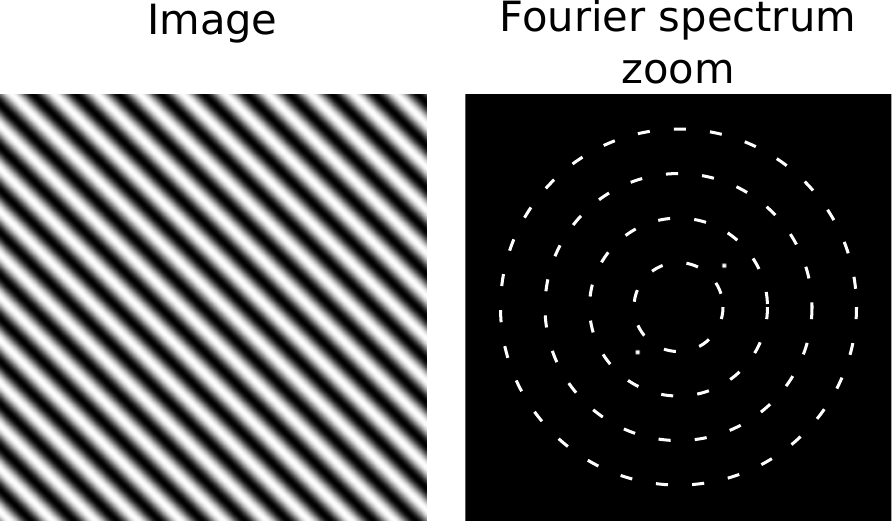}
   \caption{Example of image with a 2D plane wave and its respective magnitude Fourier spectrum. Note that there are only two points in the spectra.  Observe that the orientation defined by these two points coincides with that of the plane wave. The dashed lines are rings with fixed width used to determine the features.}
  \label{fig:fourier}
\end{figure}

\end{enumerate}

\section{Results and Discussion}\label{sec:results}

In this section we will provide a qualitative analysis of the visual properties of the mesoscopic networks studied in this work, complemented by the discussion of accuracy of the authorship attribution results obtained from experiments considering all authors and experiments considering only pairs of authors. Finally, we provide a principal component analysis in order to visualize the data clusterization.

\subsection{Image analysis}\label{imageanalysis}

\begin{figure*}[!htpb]
  \centering
    \includegraphics[width=.8\textwidth]{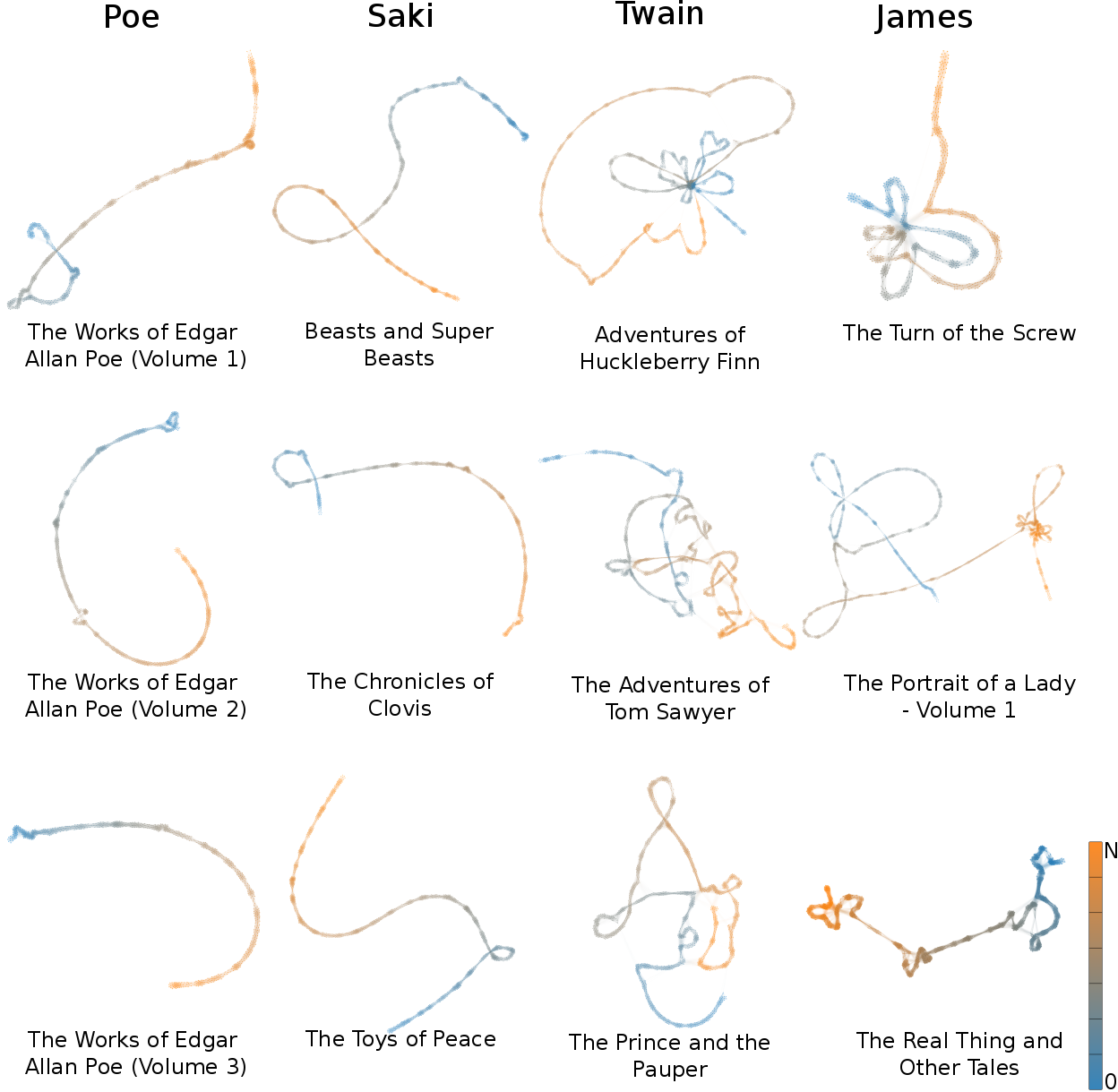}
   \caption{The discussed mesoscopic networks.}
  \label{fig:networks}
\end{figure*}

So as to illustrate some of the visual specificities present in the constructed networks in this paper, we used a visualization method to generate three networks from each of four different authors. These were: Edgar Allan Poe, Saki, Mark Twain and Henry James. These networks are presented in Figure~\ref{fig:networks}. In order to better visualize the narrative flow, color was added to the  images. Several of the loops that appear in the images are a consequence of the network visualization. These loops could have been removed manually, but we chose not to interfere in any step of the process.

The obtained images suggest a kind of \emph{visual network} whose edges are narrow strings while the nodes correspond to intersections appearing along these strings.  Observe that the string extremities also define nodes. Pieces of the same string initiating and terminating in a node are henceforth called \emph{visual loops}. Nodes with several connections are called \emph{visual hubs}.  Two types of these hubs can be identified in Figure~\ref{fig:networks}: \emph{tight hubs}, as seen in Huckleberry Finn's network; and \emph{loose hubs}, found in The Turn of the Screw.  

The images' most noticeable feature is clearly the absence or presence of visual nodes in the networks.  Both Saki and Poe have networks without many nodes. This can easily be explained since them both favor collections of  unrelated short stories. Nonetheless, there are some small differences between the aforementioned networks; Poe's networks depict some small scale loops that are absent in Saki's networks. We hypothesize that, although these differences are visually minor, they will have a notable effect on the respective mesoscopic network measurements. 

Henry James' The Real Thing and Other Tales also present small scale loops, but in this case they are more pronounced. Contrariwise, in the other books from James, specially in The Turn of The Screw, the presence of longer loops is more prevalent. We attribute this effect to the book's plot, where almost all the story consists of the narration of a single character and her interactions with only three other people inside a house, using therefore similar vocabulary throughout the book. We also suspect that by being a suspense/horror story the narrative keeps returning to the same place in order to build up suspense.

Twain's books also have their own peculiarities. His books' networks are much more convoluted than the other authors', having many more long loops. This is indicative that he had a preference for re-using specific patterns throughout his novels. These patterns could possibly be attributed to either characters' speech patterns or situational vocabulary. The former are easily seen in Huckleberry Finn's network, organized around a tight visual hub. In this novel, Twain tried to represent the phonetic differences of a slave's speech pattern. Therefore, whenever Jim (a recently freed slave) interacts with other characters, several distinctive words are used, causing it to have a very typical tf-idf distribution, leading to the tight hub.

It is interesting to notice that both Huckleberry Finn and The Turn of the Screw have many long loops and also present a central hub, but there are clear differences. We suppose that these differences are due to distinct underlying reasons. In Huckleberry Finn, the long loops appear to be caused by a single character distinctive vocabulary. On the other hand, The Turn of the Screw is a book that takes place mostly in the same location, with few different characters, and therefore shares similar situations throughout the book, making these long loops more spread out.

%\begin{figure*}[!htpb]
%  \centering
%    \includegraphics[width=1.\textwidth]{example.png}
%   \caption{Example of networks obtained from two distinct authors. Note that the elongation of the images are clearly different when we compare the two first images with the third image. This measurement is 2.50, 1.37, and 9.76, for the networks created from two first books of Mark Twain, and Edgar Alan Poe, respectively.}
%  \label{fig:images}
%\end{figure*}

%\Red{Comparar algumas imagens (umas 3), colocar uma ou 2 features mostrando a característica da imagem refletida na feature. Isso seria um exemplo que separa.}

\subsection{Authorship Attribution}\label{allres}
In order to evaluate the capability of the proposed method to classify books according to their respective authors, we calculated all the image features, IF henceforth, described in Section~\ref{sec:imageAnalysis}. These features were selected and ordered according to SVM's (Support Vector Machine) attribute selection~\cite{Guyon2002}. Furthermore, we used expectation maximization (EM)~\cite{lloyd1982least} and KMeans~\cite{dempster1977maximum}, which are unsupervised classifier algorithms.  The number of classes was set to 10 in order to reflect the number of authors. In order to find the best number of features, we tested the $n$ first features in the ranking by varying $n$ from 1 to $\mathcal{F}$, where $\mathcal{F}$ is the total number of features (see Table~\ref{results}). The best classification accuracy, $54\%$, was achieved using $n=14$. This result supports the hypothesis that there are visual differences among the visualizations of the mesoscopic networks, which are captured by the image features.
%SVM Attibute selection: I. Guyon, J. Weston, S. Barnhill, V. Vapnik (2002). Gene selection for cancer classification using support vector machines. Machine Learning. 46:389-422.

An alternative method to classify these texts is to use features obtained from the complex network measurements, or network features (NF). The selected measurements were presented in Section~\ref{sec:networkAnalysis}. We employed the same machine learning methodology that was applied to the image features and obtained an accuracy of $54\%$ for $n=15$. We also combined all features (IF~+~NF) and classified the texts using the same methodology. In this case, a better accuracy rate was achieved: $58\%$, with $n=19$. 

\begin{table}[]
\centering
\caption{Classification accuracy and number of used features considering different measurements.}
\label{results}
\begin{tabular}{c|c|c|c|c|}
\cline{2-5}
                            & \multicolumn{2}{c|}{EM} & \multicolumn{2}{c|}{KMeans} \\ \cline{2-5} 
                            & Accuracy   & Features   & Accuracy     & Features     \\ \hline
\multicolumn{1}{|c|}{IF}    & 54\%       & 14         & 48\%         & 13           \\ \hline
\multicolumn{1}{|c|}{NF}    & 54\%       & 15         & 50\%         & 3            \\ \hline
\multicolumn{1}{|c|}{IF+NF} & 58\%       & 19         & 54\%         & 34           \\ \hline
\end{tabular}
\end{table}

\subsection{Pairwise Authorship Attribution}
\begin{figure*}[ht!]
    \centering
    \subfigure[IF]{\includegraphics[width=0.32\textwidth]{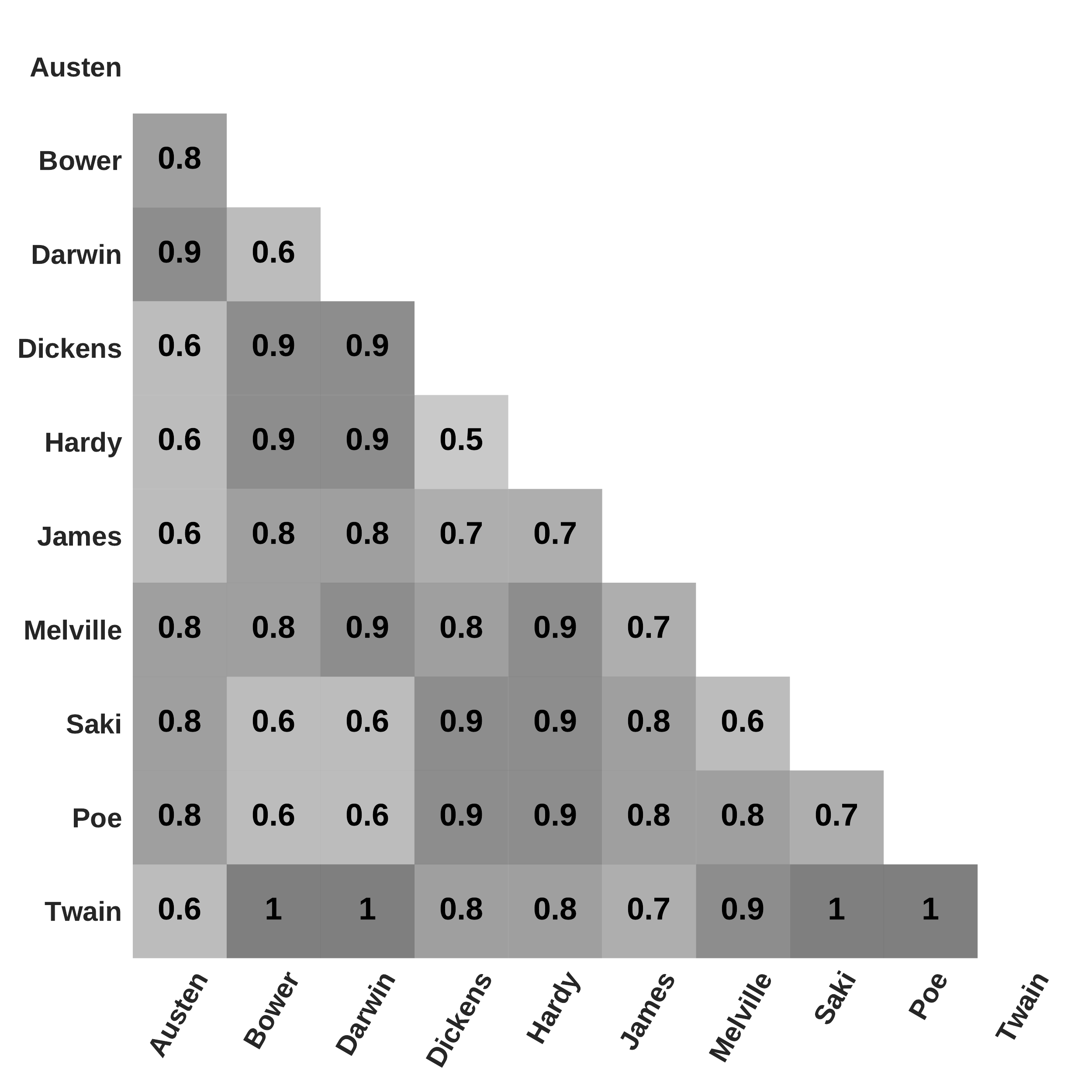}}
    \subfigure[NF]{\includegraphics[width=0.32\textwidth]{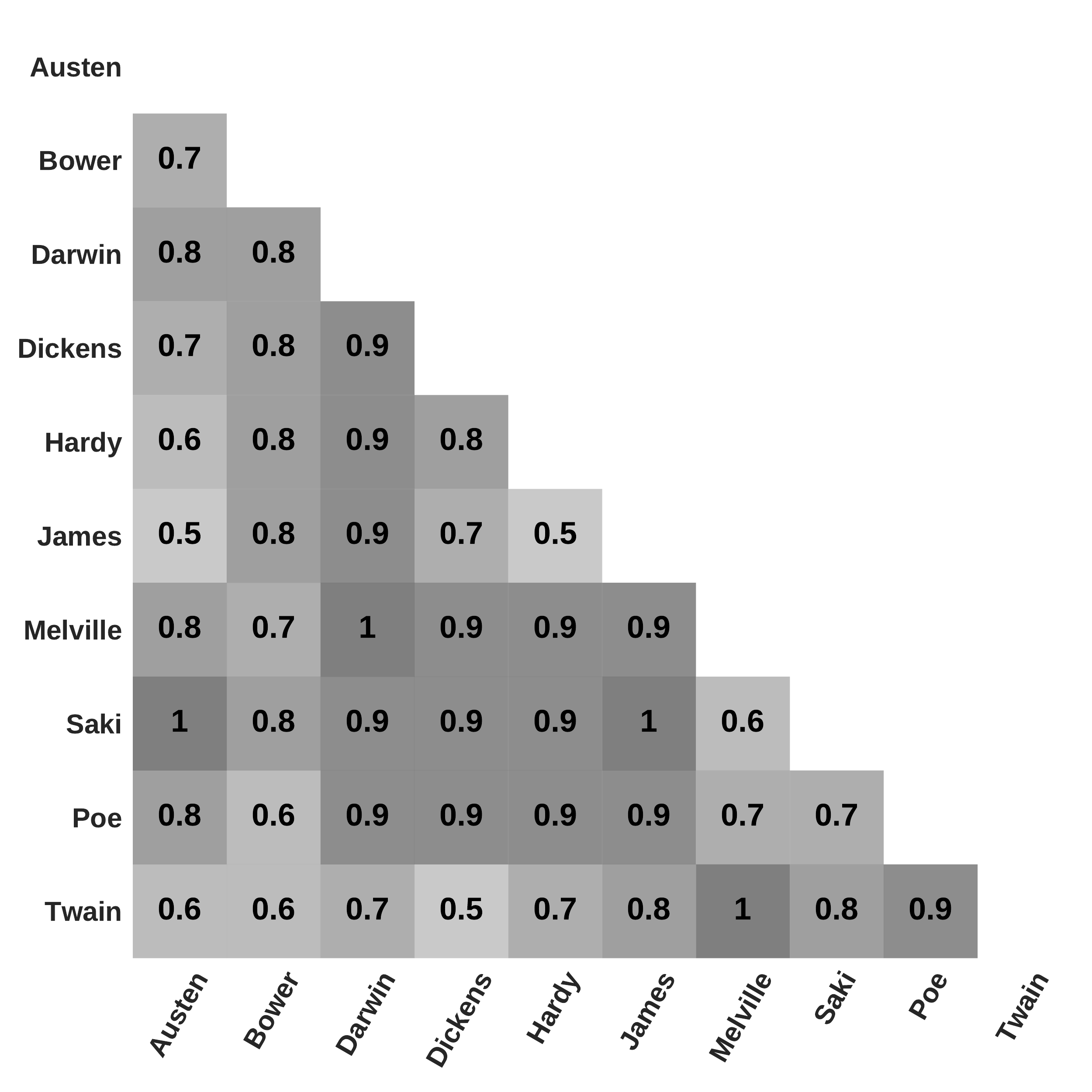}}
    \subfigure[IF + NF]{\includegraphics[width=0.32\textwidth]{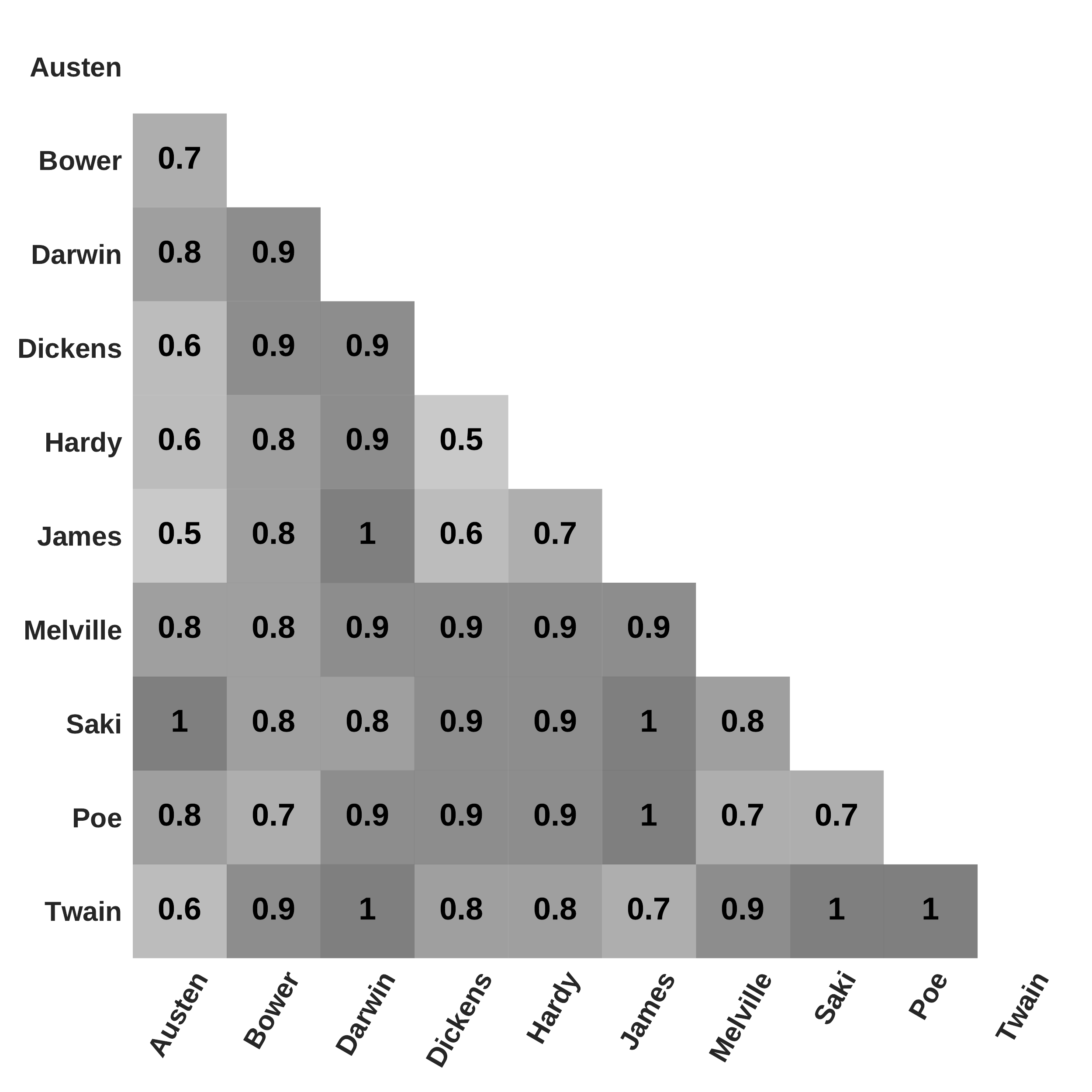}}
    \caption{Accuracy rates (from 0 to 1) in the pairwise classification using only image features (a) - IF, only network features (b) - NF, and the combination of the two strategies (c) - IF + NF. Note that, in general, the accuracies increase when both strategies are combined and more pairs are classified with accuracy of 1.}\label{pairwise}
\end{figure*}

In Figure~\ref{pairwise} we have the results for pairwise classifications, where only two authors are considered for each experiment. We utilize the classifiers with best accuracy from the previous sub section. We can see that, as expected, the results that consider both image and network features are better than those that consider only one type of feature. 

Note that the results reflect our analysis in subsection \ref{imageanalysis}: while the result for the comparison between Poe and Saki, and James and Twain yielded an accuracy of 70\%, all other results have an 100\% accuracy. This is a clear indication that the features selected reflect our intuitions and that it is even possible to discriminate between similar networks, albeit improvements are desirable.

\subsection{PCA}

\begin{figure*}[ht!]
    \centering
    \subfigure[]{\includegraphics[width=0.49\textwidth]{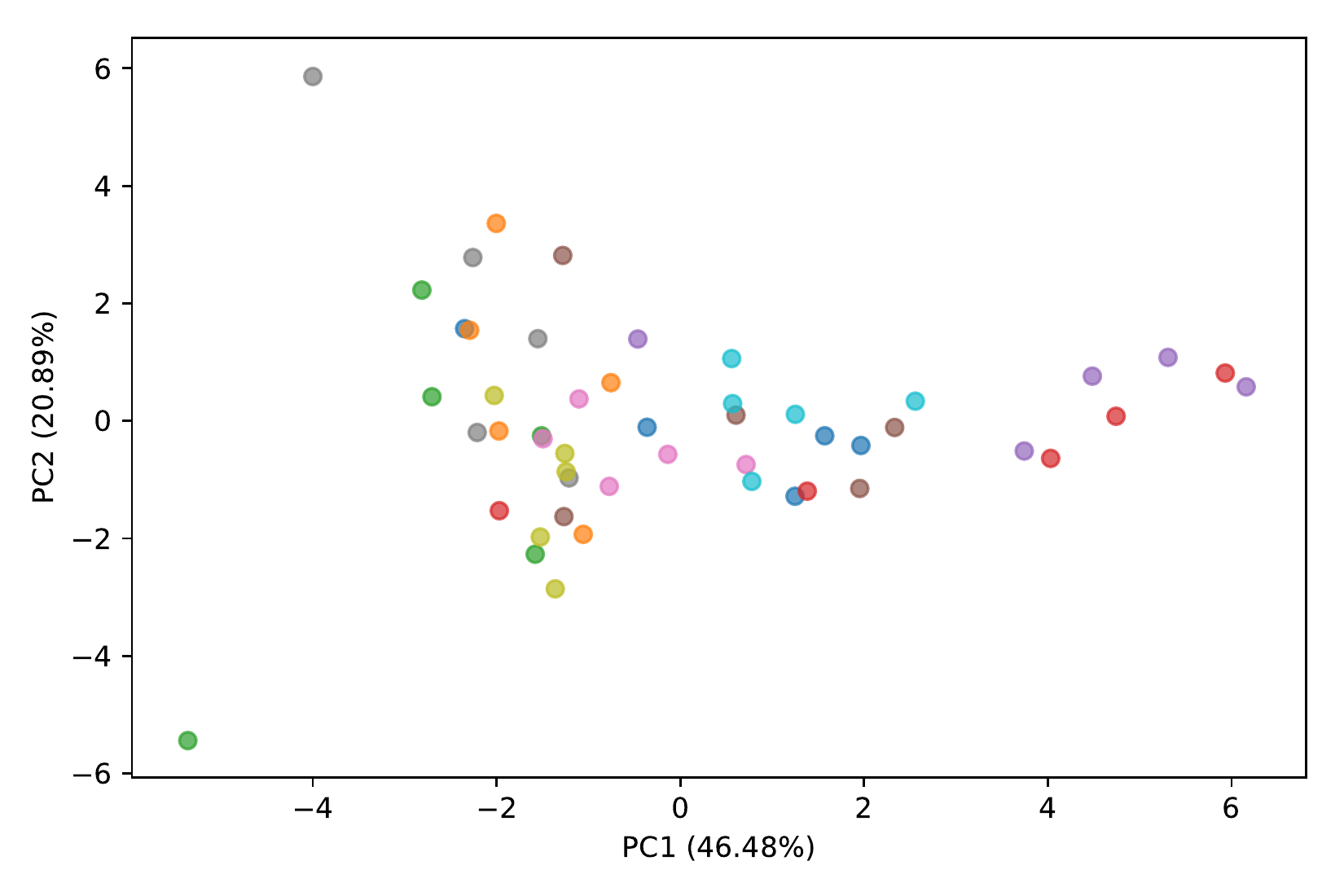}}
    \subfigure[]{\includegraphics[width=0.49\textwidth]{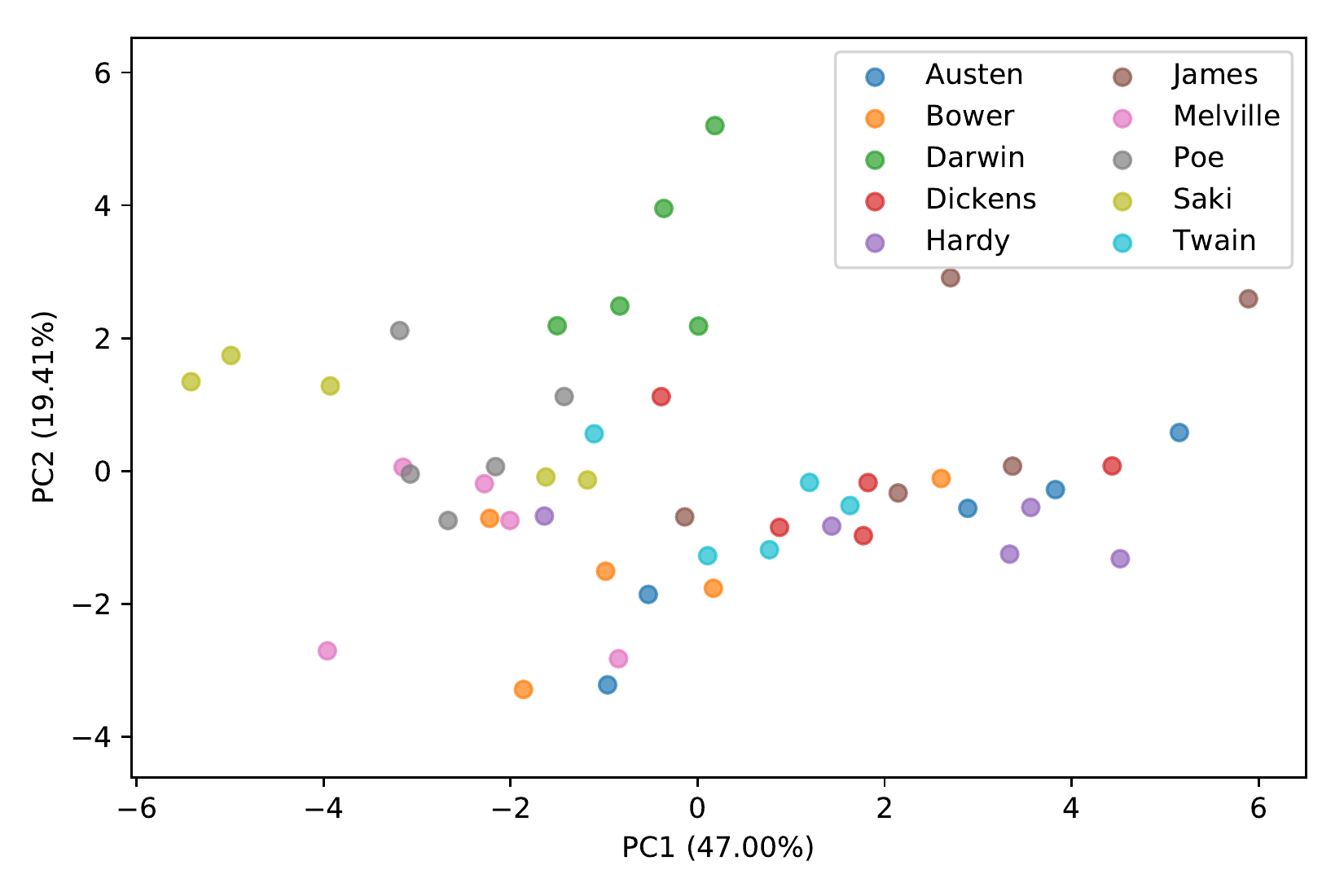}}
    \caption{PCA projections of the generated networks using only image (a) or network  (b) features.}\label{pca}
\end{figure*}

\begin{figure*}[!htpb]
  \centering
    \includegraphics[width=0.8\textwidth]{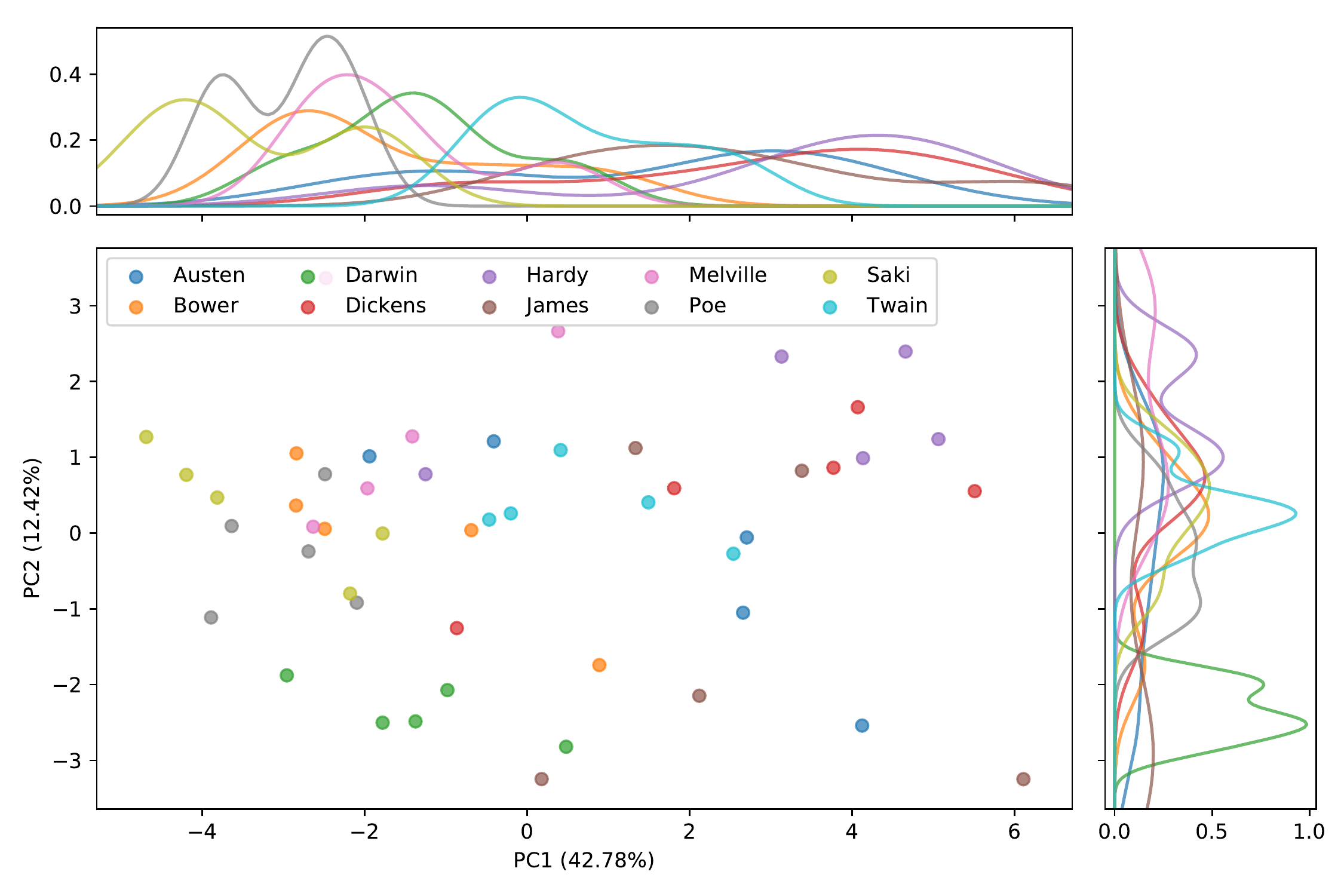}
   \caption{PCA projections of the generated networks using both image and network features.}
  \label{fig:all}
\end{figure*}

\begin{figure}[ht!]
    \centering
   \subfigure[PC1]{\includegraphics[width=0.49\textwidth]{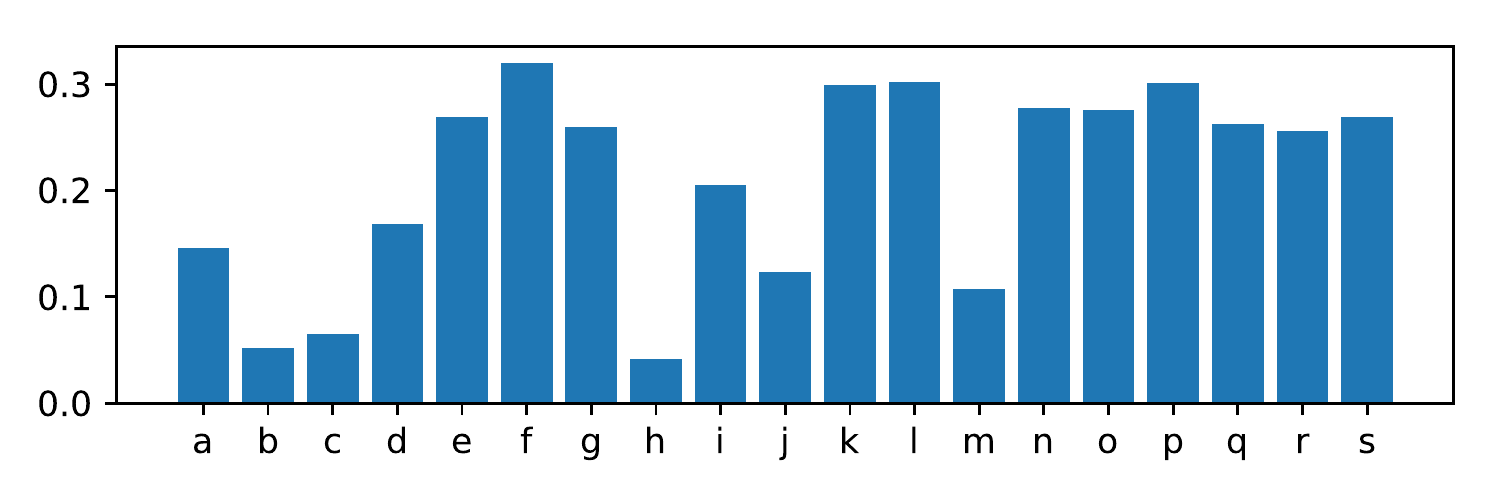}}
    \subfigure[PC2]{\includegraphics[width=0.49\textwidth]{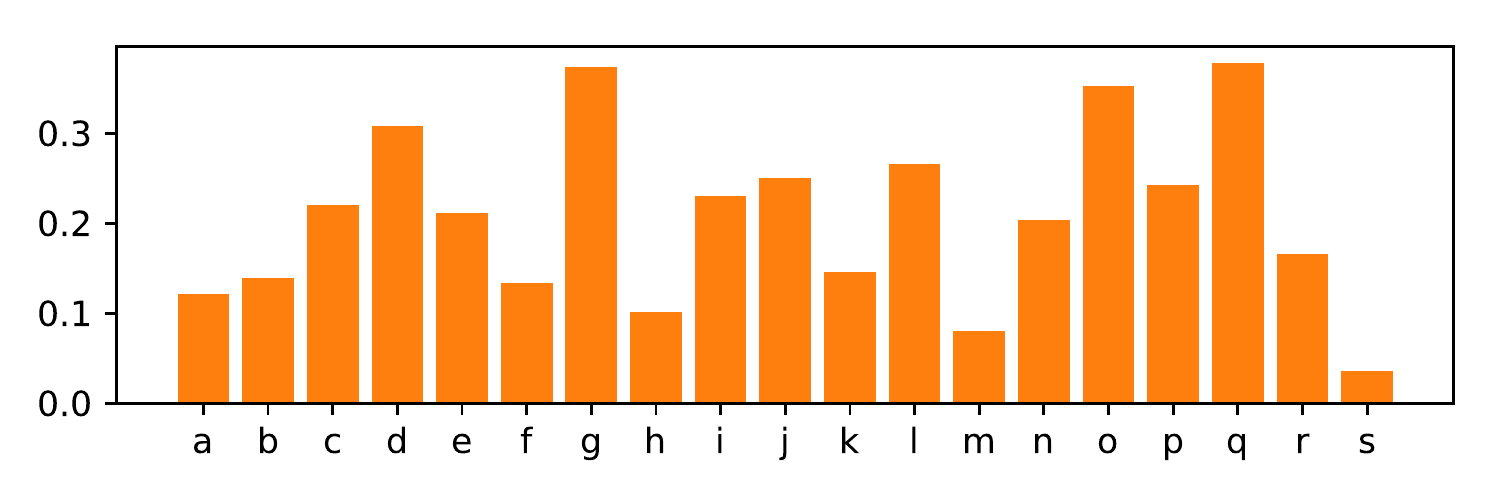}}
    \caption{Weights (in absolute values) of PCA (shown in Figure~\ref{fig:all}) computed from the IF+NF measurements. Item (a) and (b) represents the first and second principal components, respectively. The considered measurements are: a~-~entropy of Fourier spectrum (ring with ratio 30); b~-~skewness of backbone symmetry ($h = 4$) c~-~standard deviation of backbone symmetry ($h = 3$); d~-~average of backbone symmetry ($h = 3$); e~-~skewness of clustering coefficient; f~-~entropy of Fourier spectrum (ratio 120); g~-~standard deviation  of Fourier spectrum (ratio 105); h~-~entropy of Fourier spectrum (ring with ratio 90); i~-~average knn; j~-~standard deviation of backbone symmetry ($h = 2$); k~-~lacunarity (ratio 31); l~-~average merged symmetry ($h = 3$); m~-~elongation; n~-~skewness of merged symmetry ($h = 4$); o~-~average of Fourier spectrum (ring with ratio 45); p~-~average merged symmetry ($h = 4$); q~-~average of Fourier spectrum (ring with ratio 105); r~-~skewness of degree, and s~-~standard deviation of accessibility ($h = 2$).}
    \label{pcaWALL}
\end{figure}

Finally, PCA was used to check if our image and network features properly represent the desired characteristics. Fourteen features were selected according to SVM feature selector. We applied PCA to the image and network features separately and achieved reasonable separation, specially with network features, as shown in Figure ~\ref{pca}.  Network and image features were combined achieving a better separation, as depicted in Figure~\ref{fig:all}. According to this figure, it is clear that our features are capable of capturing some stylistic choices of the authors. For example, Poe and Saki's generated networks have similar structures and are fairly close to one another, but the features are distinctive enough to produce very little actual overlap. It is also interesting to note that the PCA weights in Figure~\ref{pcaWALL} show that both image and network features have strong contribution to the separation.

%\begin{table}[ht]
%\centering
%\caption{Classification accuracy and number of used features from the different measurement approaches.}
%\label{results}
%\begin{tabular}{|l|l|l|l|}
%\cline{1-3}
%      & Features & Accuracy \\ \cline{1-3}
%IF    & 16       & $50\%$   \\ \cline{1-3}
%NF    & 5        & $52\%$   \\ \cline{1-3}
%IF+NF & 6        & $60\%$   \\ \cline{1-3}
%\end{tabular}
%\end{table}

%\begin{figure*}[!htpb]
%  \centering
%    \includegraphics[width=0.8\textwidth]{PCA1_Final_Images.pdf}
%   \caption{Image features.}
%  \label{fig:images}
%\end{figure*}

%\begin{figure*}[!htpb]
%  \centering
%    \includegraphics[width=0.8\textwidth]%{PCA1_Final_Networks.pdf}
%   \caption{Network features.}
% \label{fig:images}
%\end{figure*}

%\begin{figure}[ht!]
%    \centering
%    \subfigure[PC1]{\includegraphics[width=0.49\textwidth]{PCA1_Final_Images_weights1Final.pdf}}
%    \subfigure[PC2]{\includegraphics[width=0.49\textwidth]{PCA1_Final_Images_weights2Final.pdf}}
%    \caption{PCA weights (IMAGES) \Red{Thales, testei novamente as features e tinha ima outra quantidade de features que dava o mesmo aceto, antes era 20 agora sao 16 (como para os demais, coloquei a menor quantodade) Obs: Não fiz as figuras finais para não perder tempo, depois coloco certo as que manteremos.}}\label{pcaWI}
%\end{figure}

%\begin{figure*}[!htpb]
%  \centering
%    \includegraphics[width=0.8\textwidth]{LDA1_Final_All.pdf}
%   \caption{LDA All features.}
%  \label{fig:LDAimages}
%\end{figure*}

\section{Conclusions}\label{sec:conclusions}

In this work we presented a distinct manner to extract meaningful features from mesoscopic networks. Our approach involves the application of a comprehensive set of image analysis techniques ranging from simpler approaches such as area and perimeter, to more sophisticated methodologies such as Fourier spectra and lacunarity.  More specifically, our results indicate that quantitative visual analysis of these networks yielded results as good as those obtained using only topological features. Moreover, the combination of both types of features improves the attribution results.  

Possible extensions of the proposed work include the consideration of other types of texts and literary periods, as well as to other types of data, such as those derived from time series.

\section*{Acknowledgments}

H.F.A. and T.S.L. thank CAPES for financial support. V.Q.M. and D.R.A. acknowledge financial support from S\~ao Paulo Research Foundation (FAPESP) (grant no. 15/05676-8, 16/19069-9). L.d.F.C. is grateful to CNPq (Brazil) (grant no.  307333/2013-2), FAPESP (grant no.  11/50761-2), and NAP-PRP-USP for sponsorship. 

\newpage

\ \\ 

\newpage

\ \\ 

\newpage

\bibliography{references}

\begin{thebibliography}{38}
\expandafter\ifx\csname natexlab\endcsname\relax\def\natexlab#1{#1}\fi
\expandafter\ifx\csname bibnamefont\endcsname\relax
  \def\bibnamefont#1{#1}\fi
\expandafter\ifx\csname bibfnamefont\endcsname\relax
  \def\bibfnamefont#1{#1}\fi
\expandafter\ifx\csname citenamefont\endcsname\relax
  \def\citenamefont#1{#1}\fi
\expandafter\ifx\csname url\endcsname\relax
  \def\url#1{\texttt{#1}}\fi
\expandafter\ifx\csname urlprefix\endcsname\relax\def\urlprefix{URL }\fi
\providecommand{\bibinfo}[2]{#2}
\providecommand{\eprint}[2][]{\url{#2}}

\bibitem[{\citenamefont{Stamatatos}(2009)}]{Stamatatos}
\bibinfo{author}{\bibfnamefont{E.}~\bibnamefont{Stamatatos}},
  \bibinfo{journal}{Journal of the American Society for Information Science and
  Technology.} \textbf{\bibinfo{volume}{60}}, \bibinfo{pages}{538}
  (\bibinfo{year}{2009}), ISSN \bibinfo{issn}{1532-2882}.

\bibitem[{\citenamefont{Juola}(2006)}]{Juola:2006}
\bibinfo{author}{\bibfnamefont{P.}~\bibnamefont{Juola}},
  \bibinfo{journal}{Foundations and Trends in Information Retrieval}
  \textbf{\bibinfo{volume}{1}}, \bibinfo{pages}{233} (\bibinfo{year}{2006}),
  ISSN \bibinfo{issn}{1554-0669}.

\bibitem[{\citenamefont{Grieve}(2007)}]{grieve2007}
\bibinfo{author}{\bibfnamefont{J.}~\bibnamefont{Grieve}},
  \bibinfo{journal}{Literary and Linguistic Computing}
  \textbf{\bibinfo{volume}{22}}, \bibinfo{pages}{251} (\bibinfo{year}{2007}).

\bibitem[{\citenamefont{Koppel et~al.}(2009)\citenamefont{Koppel, Schler, and
  Argamon}}]{Koppel:2009}
\bibinfo{author}{\bibfnamefont{M.}~\bibnamefont{Koppel}},
  \bibinfo{author}{\bibfnamefont{J.}~\bibnamefont{Schler}}, \bibnamefont{and}
  \bibinfo{author}{\bibfnamefont{S.}~\bibnamefont{Argamon}},
  \bibinfo{journal}{Journal of the American Society for Information Science and
  Technology.} \textbf{\bibinfo{volume}{60}}, \bibinfo{pages}{9}
  (\bibinfo{year}{2009}), ISSN \bibinfo{issn}{1532-2882}.

\bibitem[{\citenamefont{Lahiri and Mihalcea}(2013)}]{Lahiri}
\bibinfo{author}{\bibfnamefont{S.}~\bibnamefont{Lahiri}} \bibnamefont{and}
  \bibinfo{author}{\bibfnamefont{R.}~\bibnamefont{Mihalcea}},
  \bibinfo{journal}{arXiv:1311.2978}  (\bibinfo{year}{2013}).

\bibitem[{\citenamefont{Marinho et~al.}(2016)\citenamefont{Marinho, Hirst, and
  Amancio}}]{Marinho2016BRACIS}
\bibinfo{author}{\bibfnamefont{V.~Q.} \bibnamefont{Marinho}},
  \bibinfo{author}{\bibfnamefont{G.}~\bibnamefont{Hirst}}, \bibnamefont{and}
  \bibinfo{author}{\bibfnamefont{D.~R.} \bibnamefont{Amancio}}, in
  \emph{\bibinfo{booktitle}{Proceedings of the 5th Brazilian Conference on
  Intelligent Systems (BRACIS)}} (\bibinfo{address}{Recife, Brazil},
  \bibinfo{year}{2016}).

\bibitem[{\citenamefont{Amancio}(2015)}]{1742-5468-2015-3-P03005}
\bibinfo{author}{\bibfnamefont{D.~R.} \bibnamefont{Amancio}},
  \bibinfo{journal}{Journal of Statistical Mechanics: Theory and Experiment}
  \textbf{\bibinfo{volume}{2015}}, \bibinfo{pages}{P03005}
  (\bibinfo{year}{2015}).

\bibitem[{\citenamefont{Segarra et~al.}(2015)\citenamefont{Segarra, Eisen, and
  Ribeiro}}]{segarra2015authorship}
\bibinfo{author}{\bibfnamefont{S.}~\bibnamefont{Segarra}},
  \bibinfo{author}{\bibfnamefont{M.}~\bibnamefont{Eisen}}, \bibnamefont{and}
  \bibinfo{author}{\bibfnamefont{A.}~\bibnamefont{Ribeiro}},
  \bibinfo{journal}{IEEE Transactions on Signal Processing}
  \textbf{\bibinfo{volume}{63}}, \bibinfo{pages}{5464} (\bibinfo{year}{2015}).

\bibitem[{\citenamefont{Amancio
  et~al.}(2012{\natexlab{a}})\citenamefont{Amancio, {Oliveira~Jr.}, and
  Costa}}]{interplay}
\bibinfo{author}{\bibfnamefont{D.~R.} \bibnamefont{Amancio}},
  \bibinfo{author}{\bibfnamefont{O.~N.} \bibnamefont{{Oliveira~Jr.}}},
  \bibnamefont{and} \bibinfo{author}{\bibfnamefont{L.~F.} \bibnamefont{Costa}},
  \bibinfo{journal}{Physica A: Statistical Mechanics and its Applications}
  \textbf{\bibinfo{volume}{391}}, \bibinfo{pages}{4406}
  (\bibinfo{year}{2012}{\natexlab{a}}).

\bibitem[{\citenamefont{Mehri et~al.}(2012)\citenamefont{Mehri, Darooneh, and
  Shariati}}]{MEHRI20122429}
\bibinfo{author}{\bibfnamefont{A.}~\bibnamefont{Mehri}},
  \bibinfo{author}{\bibfnamefont{A.~H.} \bibnamefont{Darooneh}},
  \bibnamefont{and} \bibinfo{author}{\bibfnamefont{A.}~\bibnamefont{Shariati}},
  \bibinfo{journal}{Physica A: Statistical Mechanics and its Applications}
  \textbf{\bibinfo{volume}{391}}, \bibinfo{pages}{2429 }
  (\bibinfo{year}{2012}).

\bibitem[{\citenamefont{Ferrer~i Cancho et~al.}(2004)\citenamefont{Ferrer~i
  Cancho, Sol\'e, and K\"ohler}}]{Cancho2004patterns}
\bibinfo{author}{\bibfnamefont{R.}~\bibnamefont{Ferrer~i Cancho}},
  \bibinfo{author}{\bibfnamefont{R.~V.} \bibnamefont{Sol\'e}},
  \bibnamefont{and} \bibinfo{author}{\bibfnamefont{R.}~\bibnamefont{K\"ohler}},
  \bibinfo{journal}{Phys. Rev. E} \textbf{\bibinfo{volume}{69}},
  \bibinfo{pages}{051915} (\bibinfo{year}{2004}).

\bibitem[{\citenamefont{Liu}(2009)}]{semantic}
\bibinfo{author}{\bibfnamefont{H.}~\bibnamefont{Liu}},
  \bibinfo{journal}{Chinese Science Bulletin} \textbf{\bibinfo{volume}{54}},
  \bibinfo{pages}{2781} (\bibinfo{year}{2009}), ISSN \bibinfo{issn}{1001-6538}.

\bibitem[{\citenamefont{Amancio et~al.}(2013)\citenamefont{Amancio, Altmann,
  Rybski, Oliveira~Jr., and Costa}}]{10.1371/journal.pone.0067310}
\bibinfo{author}{\bibfnamefont{D.~R.} \bibnamefont{Amancio}},
  \bibinfo{author}{\bibfnamefont{E.~G.} \bibnamefont{Altmann}},
  \bibinfo{author}{\bibfnamefont{D.}~\bibnamefont{Rybski}},
  \bibinfo{author}{\bibfnamefont{O.~N.} \bibnamefont{Oliveira~Jr.}},
  \bibnamefont{and} \bibinfo{author}{\bibfnamefont{L.~F.} \bibnamefont{Costa}},
  \bibinfo{journal}{PLoS ONE} \textbf{\bibinfo{volume}{8}},
  \bibinfo{pages}{e67310} (\bibinfo{year}{2013}).

\bibitem[{\citenamefont{Amancio
  et~al.}(2012{\natexlab{b}})\citenamefont{Amancio, Oliveira~Jr., and
  Costa}}]{0295-5075-98-1-18002}
\bibinfo{author}{\bibfnamefont{D.~R.} \bibnamefont{Amancio}},
  \bibinfo{author}{\bibfnamefont{O.~N.} \bibnamefont{Oliveira~Jr.}},
  \bibnamefont{and} \bibinfo{author}{\bibfnamefont{L.~F.} \bibnamefont{Costa}},
  \bibinfo{journal}{EPL (Europhysics Letters)} \textbf{\bibinfo{volume}{98}},
  \bibinfo{pages}{18002} (\bibinfo{year}{2012}{\natexlab{b}}).

\bibitem[{\citenamefont{de~Arruda et~al.}(2017)\citenamefont{de~Arruda, Silva,
  Marinho, Amancio, and Costa}}]{de2017mesoscopic}
\bibinfo{author}{\bibfnamefont{H.~F.} \bibnamefont{de~Arruda}},
  \bibinfo{author}{\bibfnamefont{F.~N.} \bibnamefont{Silva}},
  \bibinfo{author}{\bibfnamefont{V.~Q.} \bibnamefont{Marinho}},
  \bibinfo{author}{\bibfnamefont{D.~R.} \bibnamefont{Amancio}},
  \bibnamefont{and} \bibinfo{author}{\bibfnamefont{L.~F.} \bibnamefont{Costa}},
  \bibinfo{journal}{Journal of Complex Networks} p. \bibinfo{pages}{cnx023}
  (\bibinfo{year}{2017}), \urlprefix\url{http://doi.org/10.1093/comnet/cnx023}.

\bibitem[{\citenamefont{Marinho et~al.}(2017)\citenamefont{Marinho, de~Arruda,
  Sinelli, Costa, and Amancio}}]{marinho2017Calligraphy}
\bibinfo{author}{\bibfnamefont{V.~Q.} \bibnamefont{Marinho}},
  \bibinfo{author}{\bibfnamefont{H.~F.} \bibnamefont{de~Arruda}},
  \bibinfo{author}{\bibfnamefont{T.}~\bibnamefont{Sinelli}},
  \bibinfo{author}{\bibfnamefont{L.~F.} \bibnamefont{Costa}}, \bibnamefont{and}
  \bibinfo{author}{\bibfnamefont{D.~R.} \bibnamefont{Amancio}}, in
  \emph{\bibinfo{booktitle}{Proceedings of TextGraphs-11: the Workshop on
  Graph-based Methods for Natural Language Processing}}
  (\bibinfo{publisher}{Association for Computational Linguistics},
  \bibinfo{address}{Vancouver, Canada}, \bibinfo{year}{2017}), pp.
  \bibinfo{pages}{1--10}.

\bibitem[{\citenamefont{Mosteller and Wallace}(1964)}]{Mosteller}
\bibinfo{author}{\bibfnamefont{F.}~\bibnamefont{Mosteller}} \bibnamefont{and}
  \bibinfo{author}{\bibfnamefont{D.~L.} \bibnamefont{Wallace}},
  \emph{\bibinfo{title}{Inference and Disputed Authorship: The Federalist
  Papers}} (\bibinfo{publisher}{Addison-Wesley}, \bibinfo{address}{Reading,
  Mass.}, \bibinfo{year}{1964}).

\bibitem[{\citenamefont{Bagnall}(2016)}]{bagnall:2016}
\bibinfo{author}{\bibfnamefont{D.}~\bibnamefont{Bagnall}}, in
  \emph{\bibinfo{booktitle}{{CLEF 2016 Evaluation Labs and Workshop -- Working
  Notes Papers, 5-8 September, {\'E}vora, Portugal}}}, edited by
  \bibinfo{editor}{\bibfnamefont{K.}~\bibnamefont{Balog}},
  \bibinfo{editor}{\bibfnamefont{L.}~\bibnamefont{Cappellato}},
  \bibinfo{editor}{\bibfnamefont{N.}~\bibnamefont{Ferro}}, \bibnamefont{and}
  \bibinfo{editor}{\bibfnamefont{C.}~\bibnamefont{Macdonald}}
  (\bibinfo{publisher}{CEUR-WS.org}, \bibinfo{year}{2016}), ISSN
  \bibinfo{issn}{1613-0073}.

\bibitem[{\citenamefont{Solorio et~al.}(2017)\citenamefont{Solorio, Rosso,
  Montes{-}y{-}G{\'{o}}mez, Shrestha, Sierra, and
  Gonz{\'{a}}lez}}]{SolorioEACL}
\bibinfo{author}{\bibfnamefont{T.}~\bibnamefont{Solorio}},
  \bibinfo{author}{\bibfnamefont{P.}~\bibnamefont{Rosso}},
  \bibinfo{author}{\bibfnamefont{M.}~\bibnamefont{Montes{-}y{-}G{\'{o}}mez}},
  \bibinfo{author}{\bibfnamefont{P.}~\bibnamefont{Shrestha}},
  \bibinfo{author}{\bibfnamefont{S.}~\bibnamefont{Sierra}}, \bibnamefont{and}
  \bibinfo{author}{\bibfnamefont{F.~A.} \bibnamefont{Gonz{\'{a}}lez}}, in
  \emph{\bibinfo{booktitle}{Proceedings of the 15th Conference of the European
  Chapter of the Association for Computational Linguistics, {EACL} 2017,
  Valencia, Spain, April 3-7, 2017, Volume 2: Short Papers}}
  (\bibinfo{year}{2017}), pp. \bibinfo{pages}{669--674}.

\bibitem[{\citenamefont{Sari et~al.}(2017)\citenamefont{Sari, Vlachos, and
  Stevenson}}]{SariEACL}
\bibinfo{author}{\bibfnamefont{Y.}~\bibnamefont{Sari}},
  \bibinfo{author}{\bibfnamefont{A.}~\bibnamefont{Vlachos}}, \bibnamefont{and}
  \bibinfo{author}{\bibfnamefont{M.}~\bibnamefont{Stevenson}}, in
  \emph{\bibinfo{booktitle}{European Chapter of the Association for
  Computational Linguistics (EACL 2017)}}, edited by
  \bibinfo{editor}{\bibfnamefont{M.}~\bibnamefont{Lapata}},
  \bibinfo{editor}{\bibfnamefont{P.}~\bibnamefont{Blunsom}}, \bibnamefont{and}
  \bibinfo{editor}{\bibfnamefont{A.}~\bibnamefont{Koller}}
  (\bibinfo{publisher}{ACL}, \bibinfo{year}{2017}), vol.~\bibinfo{volume}{2}.

\bibitem[{\citenamefont{Manning and Sch\"{u}tze}(1999)}]{Manning:1999}
\bibinfo{author}{\bibfnamefont{C.~D.} \bibnamefont{Manning}} \bibnamefont{and}
  \bibinfo{author}{\bibfnamefont{H.}~\bibnamefont{Sch\"{u}tze}},
  \emph{\bibinfo{title}{Foundations of Statistical Natural Language
  Processing}} (\bibinfo{publisher}{MIT Press}, \bibinfo{address}{Cambridge,
  MA, USA}, \bibinfo{year}{1999}), ISBN \bibinfo{isbn}{0-262-13360-1}.

\bibitem[{\citenamefont{Travençolo and Costa}(2008)}]{Travencolo2008}
\bibinfo{author}{\bibfnamefont{B.}~\bibnamefont{Travençolo}} \bibnamefont{and}
  \bibinfo{author}{\bibfnamefont{L.~F.} \bibnamefont{Costa}},
  \bibinfo{journal}{Physics Letters A} \textbf{\bibinfo{volume}{373}},
  \bibinfo{pages}{89 } (\bibinfo{year}{2008}), ISSN \bibinfo{issn}{0375-9601}.

\bibitem[{\citenamefont{Costa et~al.}(2007)\citenamefont{Costa, Rodrigues,
  Travieso, and Villas~Boas}}]{costa2007characterization}
\bibinfo{author}{\bibfnamefont{L.~F.} \bibnamefont{Costa}},
  \bibinfo{author}{\bibfnamefont{F.~A.} \bibnamefont{Rodrigues}},
  \bibinfo{author}{\bibfnamefont{G.}~\bibnamefont{Travieso}}, \bibnamefont{and}
  \bibinfo{author}{\bibfnamefont{P.~R.} \bibnamefont{Villas~Boas}},
  \bibinfo{journal}{Advances in physics} \textbf{\bibinfo{volume}{56}},
  \bibinfo{pages}{167} (\bibinfo{year}{2007}).

\bibitem[{\citenamefont{Silva et~al.}(2016{\natexlab{a}})\citenamefont{Silva,
  Comin, Peron, Rodrigues, Ye, Wilson, Hancock, and F.~Costa}}]{Silva}
\bibinfo{author}{\bibfnamefont{F.~N.} \bibnamefont{Silva}},
  \bibinfo{author}{\bibfnamefont{C.~H.} \bibnamefont{Comin}},
  \bibinfo{author}{\bibfnamefont{T.~K.} \bibnamefont{Peron}},
  \bibinfo{author}{\bibfnamefont{F.~A.} \bibnamefont{Rodrigues}},
  \bibinfo{author}{\bibfnamefont{C.}~\bibnamefont{Ye}},
  \bibinfo{author}{\bibfnamefont{R.~C.} \bibnamefont{Wilson}},
  \bibinfo{author}{\bibfnamefont{E.~R.} \bibnamefont{Hancock}},
  \bibnamefont{and} \bibinfo{author}{\bibfnamefont{L.}~\bibnamefont{F.~Costa}},
  \bibinfo{journal}{Information Science} \textbf{\bibinfo{volume}{333}},
  \bibinfo{pages}{61} (\bibinfo{year}{2016}{\natexlab{a}}), ISSN
  \bibinfo{issn}{0020-0255}.

\bibitem[{\citenamefont{Newman}(2003)}]{newman2003mixing}
\bibinfo{author}{\bibfnamefont{M.}~\bibnamefont{Newman}},
  \bibinfo{journal}{Physical Review E} \textbf{\bibinfo{volume}{67}},
  \bibinfo{pages}{026126} (\bibinfo{year}{2003}).

\bibitem[{\citenamefont{Pastor-Satorras
  et~al.}(2001)\citenamefont{Pastor-Satorras, V{\'a}zquez, and
  Vespignani}}]{pastor2001dynamical}
\bibinfo{author}{\bibfnamefont{R.}~\bibnamefont{Pastor-Satorras}},
  \bibinfo{author}{\bibfnamefont{A.}~\bibnamefont{V{\'a}zquez}},
  \bibnamefont{and}
  \bibinfo{author}{\bibfnamefont{A.}~\bibnamefont{Vespignani}},
  \bibinfo{journal}{Physical Review Letters} \textbf{\bibinfo{volume}{87}},
  \bibinfo{pages}{258701} (\bibinfo{year}{2001}).

\bibitem[{\citenamefont{Fruchterman and Reingold}(1991)}]{fruchterman1991graph}
\bibinfo{author}{\bibfnamefont{T.~M.~J.} \bibnamefont{Fruchterman}}
  \bibnamefont{and} \bibinfo{author}{\bibfnamefont{E.~M.}
  \bibnamefont{Reingold}}, \bibinfo{journal}{Software: Practice and experience}
  \textbf{\bibinfo{volume}{21}}, \bibinfo{pages}{1129} (\bibinfo{year}{1991}).

\bibitem[{\citenamefont{Silva et~al.}(2016{\natexlab{b}})\citenamefont{Silva,
  Amancio, Bardosova, Costa, and Oliveira~Jr}}]{silva2016using}
\bibinfo{author}{\bibfnamefont{F.~N.} \bibnamefont{Silva}},
  \bibinfo{author}{\bibfnamefont{D.~R.} \bibnamefont{Amancio}},
  \bibinfo{author}{\bibfnamefont{M.}~\bibnamefont{Bardosova}},
  \bibinfo{author}{\bibfnamefont{L.~d.~F.} \bibnamefont{Costa}},
  \bibnamefont{and} \bibinfo{author}{\bibfnamefont{O.~N.}
  \bibnamefont{Oliveira~Jr}}, \bibinfo{journal}{Journal of Informetrics.}
  \textbf{\bibinfo{volume}{10}}, \bibinfo{pages}{487}
  (\bibinfo{year}{2016}{\natexlab{b}}).

\bibitem[{\citenamefont{Efford}(2000)}]{delacao}
\bibinfo{author}{\bibfnamefont{N.}~\bibnamefont{Efford}},
  \emph{\bibinfo{title}{Digital Image Processing: A Practical Introduction
  using Java}} (\bibinfo{publisher}{Addison-Wesley}, \bibinfo{address}{Harlow,
  England}, \bibinfo{year}{2000}).

\bibitem[{\citenamefont{Jolliffe}(2002)}]{jolliffe2002principal}
\bibinfo{author}{\bibfnamefont{I.}~\bibnamefont{Jolliffe}},
  \emph{\bibinfo{title}{Principal component analysis}}
  (\bibinfo{publisher}{Wiley Online Library}, \bibinfo{year}{2002}).

\bibitem[{\citenamefont{Skyum}(1991)}]{SKYUM1991121}
\bibinfo{author}{\bibfnamefont{S.}~\bibnamefont{Skyum}},
  \bibinfo{journal}{Information Processing Letters}
  \textbf{\bibinfo{volume}{37}}, \bibinfo{pages}{121} (\bibinfo{year}{1991}).

\bibitem[{\citenamefont{Sklansky}(1982)}]{Sklansky}
\bibinfo{author}{\bibfnamefont{J.}~\bibnamefont{Sklansky}},
  \bibinfo{journal}{Pattern Recogn. Lett.} \textbf{\bibinfo{volume}{1}},
  \bibinfo{pages}{79} (\bibinfo{year}{1982}), ISSN \bibinfo{issn}{0167-8655}.

\bibitem[{\citenamefont{Costa and Cesar}(2000)}]{costa2000shape}
\bibinfo{author}{\bibfnamefont{L.~d.~F.} \bibnamefont{Costa}} \bibnamefont{and}
  \bibinfo{author}{\bibfnamefont{R.~M.} \bibnamefont{Cesar}},
  \emph{\bibinfo{title}{Shape Analysis and Classification: Theory and
  Practice}}, Image Processing Series (\bibinfo{publisher}{Taylor \& Francis},
  \bibinfo{year}{2000}), ISBN \bibinfo{isbn}{9780849334931}.

\bibitem[{\citenamefont{Plotnick et~al.}(1993)\citenamefont{Plotnick, Gardner,
  and O'Neill}}]{Plotnick1993}
\bibinfo{author}{\bibfnamefont{R.~E.} \bibnamefont{Plotnick}},
  \bibinfo{author}{\bibfnamefont{R.~H.} \bibnamefont{Gardner}},
  \bibnamefont{and} \bibinfo{author}{\bibfnamefont{R.~V.}
  \bibnamefont{O'Neill}}, \bibinfo{journal}{Landscape Ecology}
  \textbf{\bibinfo{volume}{8}}, \bibinfo{pages}{201} (\bibinfo{year}{1993}),
  ISSN \bibinfo{issn}{1572-9761}.

\bibitem[{\citenamefont{Rodrigues et~al.}(2005)\citenamefont{Rodrigues,
  Barbosa, and Costa}}]{PhysRevE.72.016707}
\bibinfo{author}{\bibfnamefont{E.~P.} \bibnamefont{Rodrigues}},
  \bibinfo{author}{\bibfnamefont{M.~S.} \bibnamefont{Barbosa}},
  \bibnamefont{and} \bibinfo{author}{\bibfnamefont{L.~d.~F.}
  \bibnamefont{Costa}}, \bibinfo{journal}{Phys. Rev. E}
  \textbf{\bibinfo{volume}{72}}, \bibinfo{pages}{016707}
  (\bibinfo{year}{2005}).

\bibitem[{\citenamefont{Guyon et~al.}(2002)\citenamefont{Guyon, Weston,
  Barnhill, and Vapnik}}]{Guyon2002}
\bibinfo{author}{\bibfnamefont{I.}~\bibnamefont{Guyon}},
  \bibinfo{author}{\bibfnamefont{J.}~\bibnamefont{Weston}},
  \bibinfo{author}{\bibfnamefont{S.}~\bibnamefont{Barnhill}}, \bibnamefont{and}
  \bibinfo{author}{\bibfnamefont{V.}~\bibnamefont{Vapnik}},
  \bibinfo{journal}{Machine Learning} \textbf{\bibinfo{volume}{46}},
  \bibinfo{pages}{389} (\bibinfo{year}{2002}).

\bibitem[{\citenamefont{Lloyd}(1982)}]{lloyd1982least}
\bibinfo{author}{\bibfnamefont{S.}~\bibnamefont{Lloyd}}, \bibinfo{journal}{IEEE
  transactions on information theory} \textbf{\bibinfo{volume}{28}},
  \bibinfo{pages}{129} (\bibinfo{year}{1982}).

\bibitem[{\citenamefont{Dempster et~al.}(1977)\citenamefont{Dempster, Laird,
  and Rubin}}]{dempster1977maximum}
\bibinfo{author}{\bibfnamefont{A.~P.} \bibnamefont{Dempster}},
  \bibinfo{author}{\bibfnamefont{N.~M.} \bibnamefont{Laird}}, \bibnamefont{and}
  \bibinfo{author}{\bibfnamefont{D.~B.} \bibnamefont{Rubin}},
  \bibinfo{journal}{Journal of the royal statistical society. Series B
  (methodological)} pp. \bibinfo{pages}{1--38} (\bibinfo{year}{1977}).

\end{thebibliography}
\end{document}